\documentclass[10pt,twocolumn,letterpaper]{article}

\usepackage[pagenumbers]{iccv} 

\definecolor{iccvblue}{rgb}{0.21,0.49,0.74}
\usepackage[pagebackref,breaklinks,colorlinks,allcolors=iccvblue]{hyperref}

\usepackage{amsmath}
\usepackage{graphicx}
\usepackage{booktabs}
\usepackage{xcolor}
\usepackage[dvipsnames]{xcolor}
\usepackage{subcaption,dcolumn}
\usepackage[shortlabels]{enumitem}

\usepackage{pifont}
\newcommand{\cmark}{\ding{51}}%
\newcommand{\xmark}{\ding{53}}

\def\paperID{6254} 
\def\confName{ICCV}
\def\confYear{2025}

\title{SatDINO: A Deep Dive into Self-Supervised Pretraining for Remote Sensing}

\author{Jakub Straka\\
University of West Bohemia in Pilsen\\
Technická 8, 301 00 Plzeň, Czech Republic\\
{\tt\small strakajk@kky.zcu.cz}
\and
Ivan Gruber\\
University of West Bohemia in Pilsen\\
Technická 8, 301 00 Plzeň, Czech Republic\\
{\tt\small grubiv@ntis.zcu.cz}
}

\begin{document}

\maketitle

\begin{abstract}
Self-supervised learning has emerged as a powerful tool for remote sensing, where large amounts of unlabeled data are available. In this work, we investigate the use of DINO, a contrastive self-supervised method, for pretraining on remote sensing imagery. We introduce SatDINO, a model tailored for representation learning in satellite imagery. Through extensive experiments on multiple datasets in multiple testing setups, we demonstrate that SatDINO outperforms other state-of-the-art methods based on much more common masked autoencoders (MAE) and achieves competitive results in multiple benchmarks.

We also provide a rigorous ablation study evaluating SatDINO's individual components. Finally, we propose a few novel enhancements, such as a new way to incorporate ground sample distance (GSD) encoding and adaptive view sampling. These enhancements can be used independently on our SatDINO model. Our code and trained models are available at: 
\url{https://github.com/strakaj/SatDINO}.
\end{abstract}

\section{Introduction}
Remote sensing is a field of research where it is relatively easy to acquire large amounts of unlabeled data from diverse sources. These sources can vary in resolution, ground sample distance (GSD), capture method (satellite, aerial, or UAV), and the underlying technology, such as RGB imagery, multi-spectral imaging (MSI), synthetic aperture radar (SAR), or LiDAR. 

In recent years, foundation models have become a standard tool in many machine learning tasks, including natural language processing and computer vision. The abundance of unlabeled data in remote sensing presents a significant opportunity for leveraging such models.

Foundation models are pre-trained in an unsupervised manner on large, unlabeled datasets. These models are then fine-tuned for specific downstream tasks. Their primary advantages include faster training times and improved performance on these downstream applications. In computer vision, unsupervised pretraining methods can be broadly classified into generative and contrastive approaches. Generative methods, such as masked image modeling (MIM), involve masking portions of an input image and using a decoder to reconstruct the missing parts, with the loss calculated between the original and reconstructed outputs. Masked Autoencoders (MAE~\cite{he2022masked}) are a main example of this approach and are widely adopted for remote sensing tasks. On the other hand, contrastive training methods focus on learning representations by comparing encoded outputs from different inputs, encouraging similarity between related data points and dissimilarity between unrelated ones. Common contrastive methods in remote sensing include CLIP~\cite{radford2021learning} (e.g., GeoCLIP~\cite{vivanco2024geoclip}, CRISP~\cite{huynh2024contrastive}) and MoCo~\cite{he2020momentum} (e.g., SeCo~\cite{manas2021seasonal}).

\begin{figure}[t]
     \centering
     \begin{subfigure}[b]{0.15\textwidth}
         \centering
         \includegraphics[width=\textwidth]{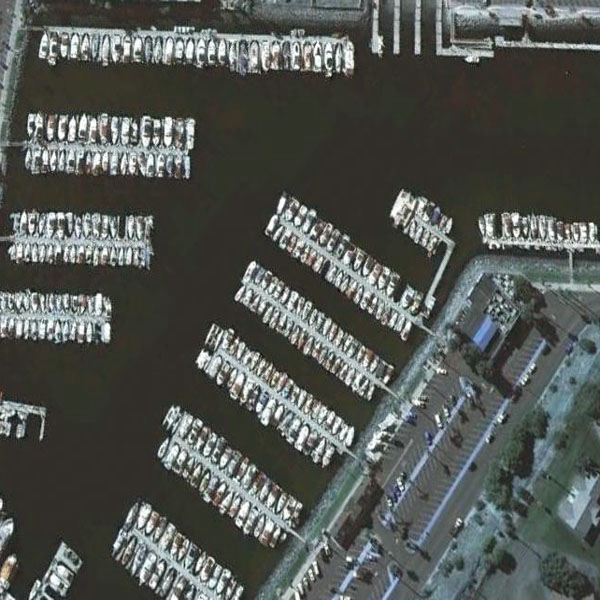}
     \end{subfigure}
     \begin{subfigure}[b]{0.15\textwidth}
         \centering
         \includegraphics[width=\textwidth]{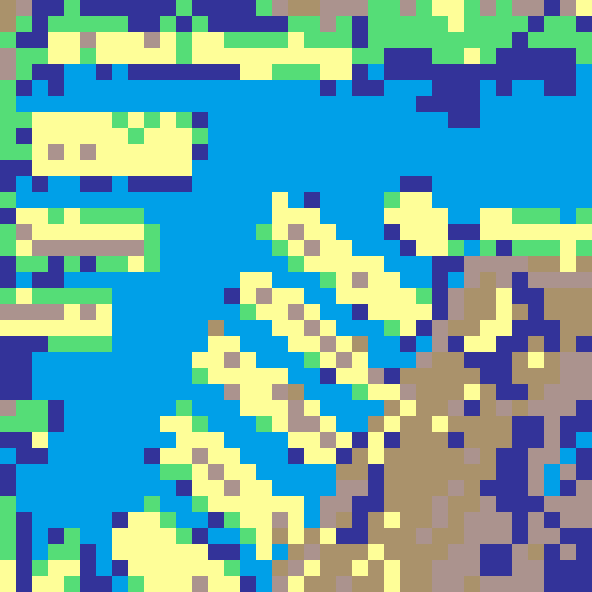}
     \end{subfigure}
    \begin{subfigure}[b]{0.15\textwidth}
         \centering
         \includegraphics[width=\textwidth]{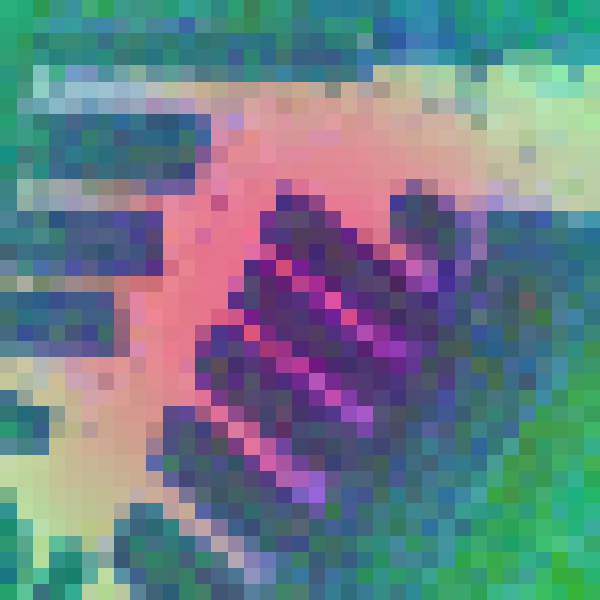}
     \end{subfigure}

     \begin{subfigure}[b]{0.15\textwidth}
         \centering
         \includegraphics[width=\textwidth]{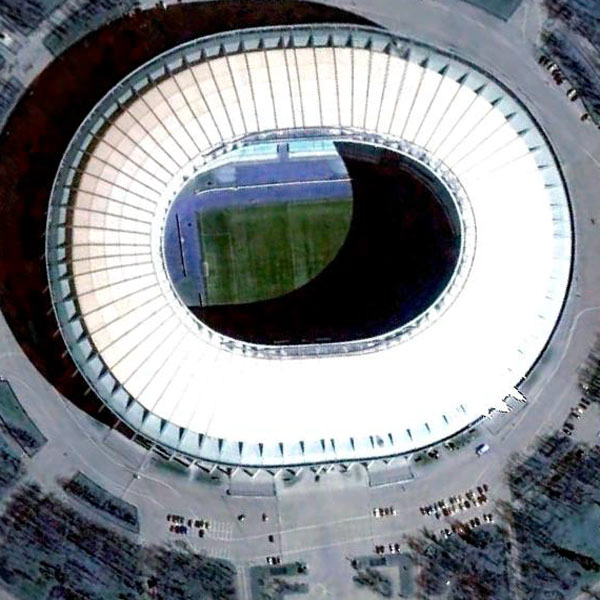}
         \caption{Original image}
     \end{subfigure}
     \begin{subfigure}[b]{0.15\textwidth}
         \centering
         \includegraphics[width=\textwidth]{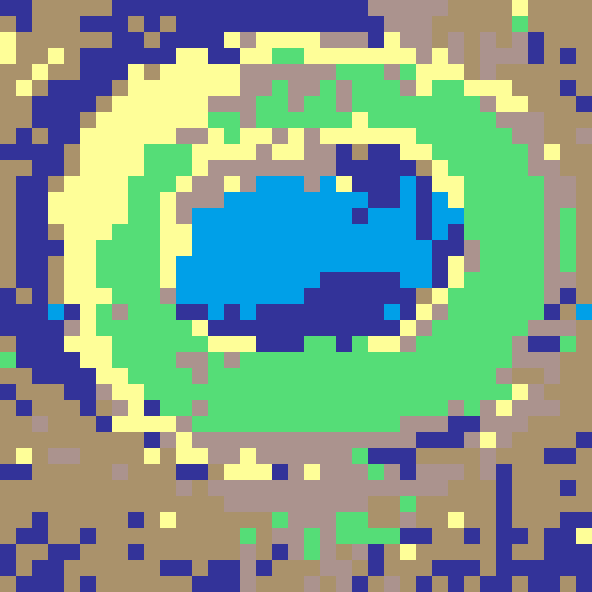}
         \caption{Attention heads}
     \end{subfigure}
    \begin{subfigure}[b]{0.15\textwidth}
         \centering
         \includegraphics[width=\textwidth]{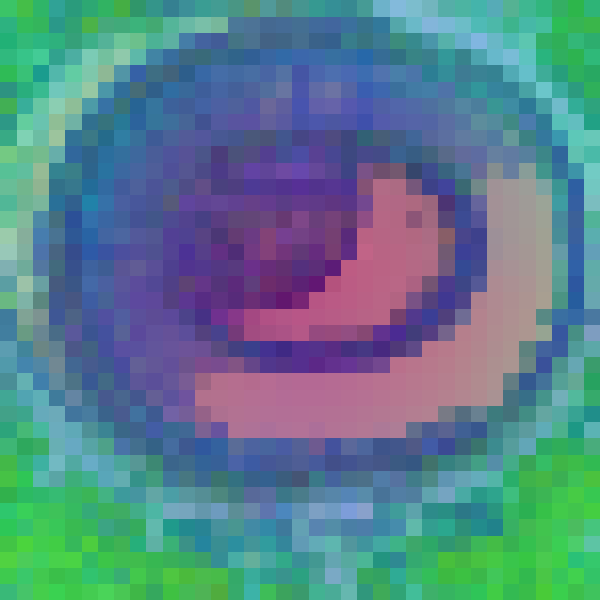}
         \caption{PCA components}
     \end{subfigure}

    \caption{\textbf{Visualization of the SatDINO model's feature representations}. (a) Original input image. (b) Attention map from the last layer, where different colors represent individual heads, and the head with the highest attention is shown for each patch. (c) PCA projection of patch embeddings.}
    \label{fig:intro}
\end{figure}

One notable recent contrastive method is DINO~\cite{caron2021emerging}, which has demonstrated strong feature representation capabilities. Except~\cite{Wanyan_2024_CVPR}, only a few works have explored DINO in the context of remote sensing. 

In this paper, we argue that DINO is better suited than MAE for pretraining on multi-scale data, as it inherently utilizes multiple scales of an image during training. We demonstrate that our proposed approach, SatDINO, learns more robust features compared to MAE-based methods. Our key contributions are:
\begin{enumerate}[(a)]
    \item We introduce \textit{uniform view sampling} and \textit{GSD encoding}, two modifications to DINO designed to better capture image scale information.
    \item We conduct a comprehensive ablation study on DINO pretraining on the functional Map of the World (fMoW-RGB)~\cite{christie2018functional} dataset.
    \item We compare SatDINO to MAE-based models, showing that DINO pretraining is more effective for multi-scale representation and produces stronger features.
\end{enumerate}


\section{Related Work}
\noindent \textbf{Generative Models.} Masked Image Modeling (MIM) is one of the most widely used pretraining tasks and serves as the foundation for the MAE framework. Many remote sensing foundation models are pre-trained using this approach.

SatMAE \cite{cong2022satmae} was among the first models trained on the remote sensing dataset fMoW. Beyond the standard MAE framework, SatMAE integrated temporal information and spectral bands alongside positional encoding. ScaleMAE \cite{reed2023scale}, another MAE-based model trained on the fMoW-RGB dataset, introduced ground sample distance (GSD) encoding combined with positional encoding to enhance scale awareness. It also utilized upsampling and Laplacian blocks to generate and combine low- and high-frequency features. Cross-Scale MAE \cite{tang2024cross}, also trained on the fMoW-RGB dataset, focused on learning multi-scale features. The model enforced consistency between outputs of an image at different scales,  encouraging the model to develop a robust understanding of scale variations.

\noindent \textbf{Contrastive models.} Contrastive learning is another popular approach for pretraining, where the outputs of two views (or modalities) are compared to encourage similarity between related data points and dissimilarity between unrelated ones. An example of this approach is the DINO framework, which aligns the representations of local and global views of the same image, encouraging consistency across these views.

In DINO-MM \cite{wang2022self}, the DINO framework is extended to align RGB data with SAR data, enabling self-supervised learning across multiple modalities. Similarly, SkySense \cite{guo2024skysense} employs contrastive training to align representations from multiple modalities.

DINO-TP and DINO-MC \cite{Wanyan_2024_CVPR} further extend the DINO framework by introducing temporal augmentations and multi-size crops, creating a more challenging pretraining task to improve feature learning.



\section{Methodology}  \label{methodology}
Our primary objective is to train the DINO model and compare its performance with MAE models~\cite{cong2022satmae, reed2023scale, tang2024cross} trained on the same dataset. This section outlines the proposed adjustments to the DINO framework, discusses alternative approaches, details the datasets used, and presents the evaluation metrics. We show detailed experiments of the individual aspects of the model in Section \ref{ablation}.

\subsection{SatDINO} 
Our model, SatDINO, builds upon the DINO framework with modifications designed for remote sensing data. We use a Vision Transformer~\cite{alexey2020image} (ViT) as the backbone and explore both small and base model variants. For the small version, we experiment with patch sizes of 16 and 8, referring to these models as SatDINO-small$_{16}$ and SatDINO-small$_{8}$, respectively. For the base version, we use only a patch size of 16, denoted as SatDINO-Base.

The core idea of DINO is to enforce alignment between local and global views of an image. Both local and global views are randomly sampled from the input image, but they differ in size. Global views cover a larger portion of the image, typically between 25\% and 100\% of the total area. Local views occupy a smaller region, ranging from 5\% to 25\% of the image. However, this aspect of the framework raises several important questions: How many views should be used? How should these views be sampled from the image? 

Additionally, remote sensing brings some differences to standard datasets such as ImageNet~\cite{deng2009imagenet}. In ImageNet, objects are typically centered and visually distinct. In contrast, satellite imagery often requires identifying features that occupy only a small portion of the image to differentiate between certain classes. Therefore the size of the local and global crops is critical, as is the strength of augmentations, which, if too aggressive, may negatively impact performance.

To address these challenges, we tested various augmentation strategies and explored multiple approaches to local view sampling.

\noindent \textbf{Augmentations.} The standard DINO framework applies augmentations such as blur, color jitter, grayscale, and solarization to local and global views. To evaluate the impact of these augmentations, we scaled their probabilities, defining two levels: \textit{soft} (scaled by 0.25) and \textit{mid} (scaled by 0.75).

Additionally, as proposed in \cite{Wanyan_2024_CVPR}, we incorporated \textit{temporal augmentation}, where multiple images of the same area, taken at different times, are used as individual views. To test this approach, we utilized two different images of the same area for global views and a third image for local views.

\noindent \textbf{Sampling of local and global views.} It is crucial for both local and global views to retain enough information to enable the teacher and student networks in DINO to produce similar output distributions. By default, DINO crops global views randomly in the range of 25\% to 100\% of the original image and local views in the range of 5\% to 25\%. After cropping, global views are resized to $224^2$, and local views are resized to $96^2$. In the context of remote sensing, cropping and resizing also alter the ground sample distance (GSD) of the image.

\begin{figure}[!htbp]
     \centering
     \begin{subfigure}[m]{0.08\textwidth}
         \centering
         \includegraphics[width=\textwidth]{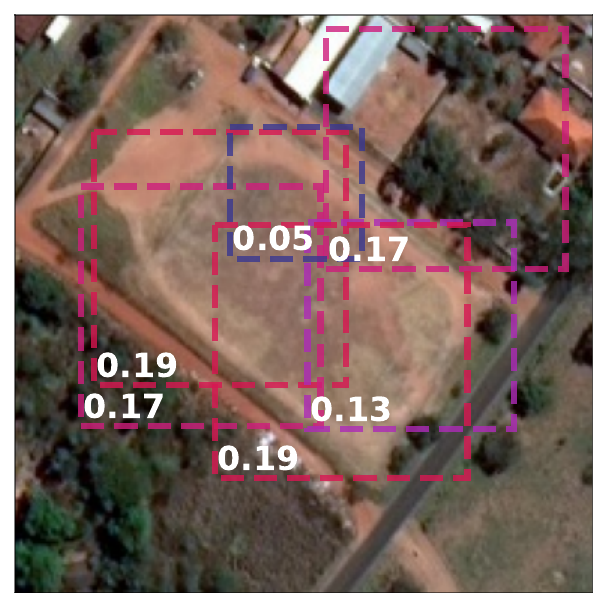}
     \end{subfigure}
     \begin{subfigure}[m]{0.39\textwidth}
         \centering
         \includegraphics[width=\textwidth]{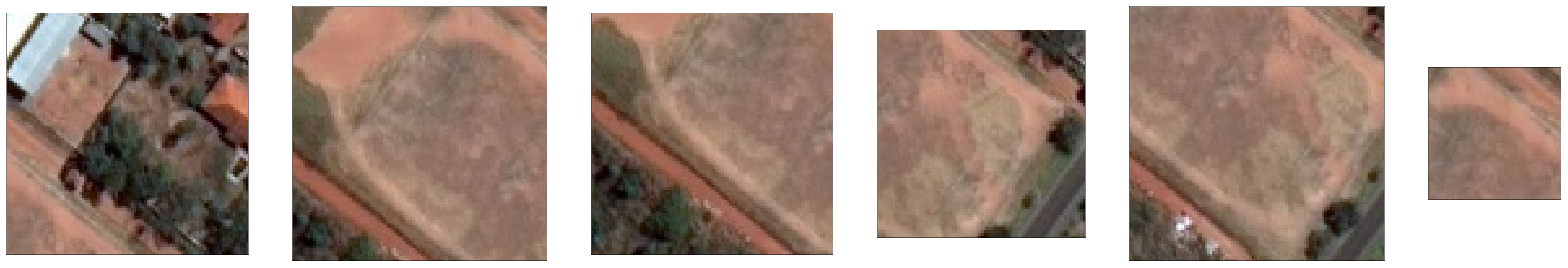}
     \end{subfigure}

     \begin{subfigure}[m]{0.08\textwidth}
         \centering
         \includegraphics[width=\textwidth]{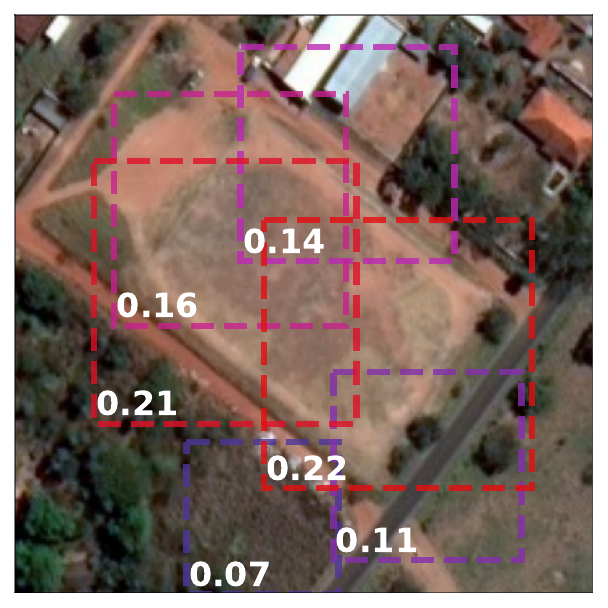}
     \end{subfigure}
     \begin{subfigure}[m]{0.39\textwidth}
         \centering
         \includegraphics[width=\textwidth]{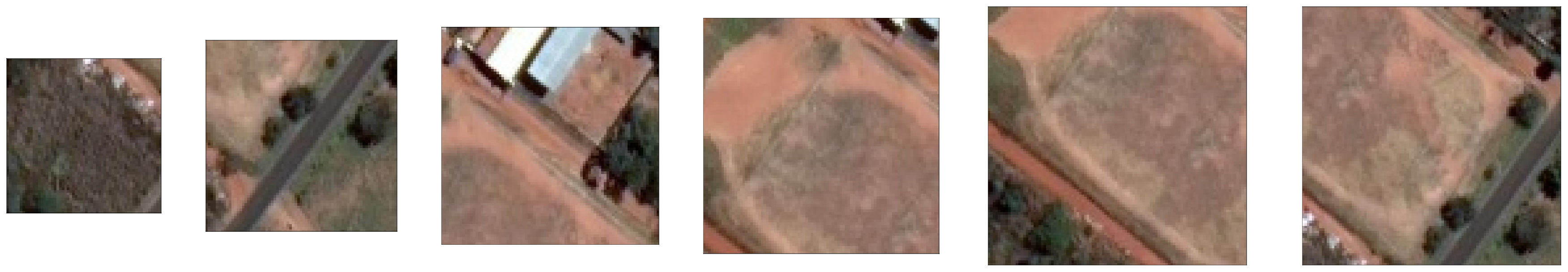}
     \end{subfigure}
    \caption{\textbf{Visualization of random and uniform sampling of the local views.} The left side shows the input image with local views outlined by dashed lines, where the number inside each square represents the area of the view. The right side shows the cropped views. In the first strategy, views are randomly sampled from the entire range. In the second strategy, the range is divided into subranges and assigned to individual views, ensuring that the first view is always the smallest and the last the largest, while still allowing for some variation in size.}
    \label{fig:views}
\end{figure}

We experimented with multiple cropping ranges and varied the number of local crops to evaluate their impact on performance. Additionally, we proposed a novel sampling strategy for local crops. We showed that training with data diverse in GSD can positively influence downstream tasks, as remote sensing datasets often include images with diverse GSD ranges (e.g., see Table~\ref{tab:classification_datasets}). Instead of sampling each local view randomly from a narrow range, we expanded the range and divided it into segments based on the number of views. Each local view was then sampled exclusively from its assigned range. This approach ensures that a broad spectrum of GSDs is represented during training. An example of random and uniform sampling for 6 local views is shown in Figure~\ref{fig:views}.

The final strategy we tested was inspired by \cite{Wanyan_2024_CVPR}. Instead of resizing all local crops to the same size, each crop was resized to a different size: $192^2$, $176^2$, $144^2$, $128^2$, $112^2$, and $96^2$.

\noindent \textbf{GSD Encoding.} The ground sample distance (GSD) provides information about the size of objects and the level of detail in satellite images. Leveraging this information can benefit model performance, for instance, methods like~\cite{reed2023scale} incorporate GSD as part of the positional encoding. However, a drawback of this approach is that the model's performance can degrade significantly if GSD information is unavailable.

To address this limitation, we propose guiding the model to estimate and be aware of GSD directly from the image itself. Specifically, we introduce a randomly initialized token, which is added alongside the class token. A linear regression layer is then added on top of this token to predict the GSD. Both the teacher and student networks include this additional token, but only the student’s outputs are optimized.

The GSD estimation uses mean squared error (MSE) as the loss function, and the overall loss is computed as a weighted combination of the DINO loss and the GSD loss: $\mathcal{L} = \mathcal{L}_{DINO} + \gamma \mathcal{L}_{GSD}$, where $\gamma$ controls the weight of the GSD loss. This approach allows the model to learn and estimate GSD directly from the image, removing the dependency on external GSD metadata during prediction.

\subsection{Method summary}
Our final SatDINO model adopts the default DINO augmentation strategy without temporal augmentations. The model retains the standard DINO cropping ranges, using global views covering 25\% to 100\% of the image and local views ranging from 5\% to 25\% while applying our uniform sampling for local crops to ensure diverse GSDs during training. Additionally, we incorporate the proposed GSD encoding approach, allowing the model to estimate GSD directly from the image. This configuration was selected as it provided the best balance between robustness and performance across our evaluation tasks.

\subsection{Model Comparison}
Our objective is to compare two pretraining frameworks: DINO and MAE. Previous works, such as \cite{cong2022satmae, reed2023scale, tang2024cross}, have modified the MAE framework to better suit satellite imagery and evaluated its performance. To ensure a fair comparison, we utilize the same dataset and similar evaluation metrics as those studies.

\begin{table}[!htbp]
    \centering
    \footnotesize
    \setlength{\tabcolsep}{0.35em}
    \begin{tabular}{l|ccccc}
        \toprule
        Dataset & GSD [m]& Train & Val. &Cat.&Res.\\
        \midrule
        fMoW-RGB~\cite{fmow2018} & 0.5 & 360105 & 52860 & 62 & $224^2$ \\
        \midrule
        EuroSAT~\cite{helber2018introducing}  &  10& 21600 & 5400 &10 & $64^2$ \\
        RESISC45~\cite{Cheng_2017} &  0.2-30& 25200 & 6300 &45 & $256^2$ \\
        UC Merced~\cite{yang2010bag} &  0.3& 1680 & 420 &21 & $256^2$ \\
        WHU-RS19~\cite{Dai2011WHURS19} &  $\leq$ 0.5& 799 & 206 &19 & $600^2$ \\
        RS-C11~\cite{zhao2016feature} &  0.2& 981 & 251 &11 & $512^2$ \\
        SIRI-WHU~\cite{zhao2015dirichlet} &  2& 1920 & 480 &12 & $200^2$ \\
         \midrule
         Potsdam~\cite{isprs}   & 0.05 & 3456 & 2016 & 6 & $512^2$ \\
         Vaihingen~\cite{isprs} & 0.09 & 344 & 398 & 6 & $512^2$ \\
         LoveDA~\cite{wang2021loveda}    & 0.3 & 2522 & 1669 & 7 & $512^2$ \\
         \bottomrule
    \end{tabular}
    \caption{\textbf{Summary of classification and segmentation datasets used for training and evaluation.} GSD for fMoW-RGB is an average; the actual GSDs are in the range from 0.307 to 1.705. The number of training and validation images for Potsdam and Vaihingen are after splitting the original images into patches.}
    \label{tab:classification_datasets}
\end{table}

\begin{figure*}[!t]
    \centering
    \includegraphics[width=0.98\linewidth]{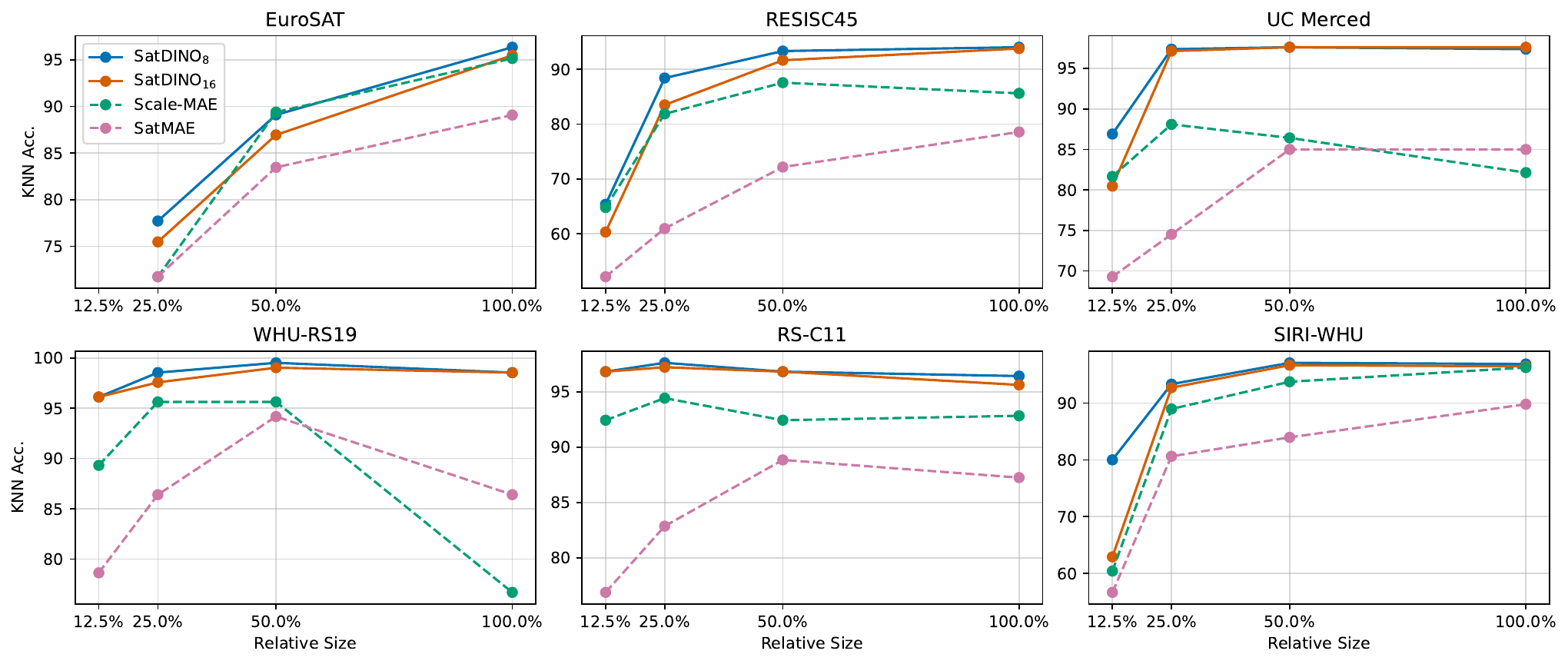}
    \caption{\textbf{kNN classification performance across multiple datasets and scales.} We compare SatDINO-Small with SatMAE and Scale-MAE by evaluating feature robustness using kNN accuracy at different scales. Our model consistently outperforms Scale-MAE across most datasets, demonstrating superior feature quality and generalization.}
    \label{fig:knn}
\end{figure*}

\noindent \textbf{Datasets.} For our primary training dataset, we use fMoW-RGB~\cite{fmow2018}. This dataset consists of 412965 images collected from various locations worldwide. The images are classified into 62 categories, span a wide range of GSDs from 0.307 to 1.705, and can be grouped into 95256 time series.

To evaluate the model's performance, we test it on a variety of downstream tasks. For classification, we utilize several datasets with diverse resolutions and GSDs, including EuroSAT~\cite{helber2018introducing}, RESISC45~\cite{Cheng_2017}, UC Merced~\cite{yang2010bag}, WHU-RS19~\cite{Dai2011WHURS19}, RS-C11~\cite{zhao2016feature}, and SIRI-WHU~\cite{zhao2015dirichlet}. Detailed information about these datasets is provided in Table \ref{tab:classification_datasets}. Some of these datasets do not include predefined train/validation splits. To ensure a fair evaluation, we divided datasets without official splits into an 80/20\% train/validation split. Our splits are publicly available for reproducibility.

For segmentation, we use the LoveDA~\cite{wang2021loveda}, Potsdam~\cite{isprs}, and Vaihingen~\cite{isprs} datasets. Both datasets are preprocessed and split using the MMSegmentation framework (MMSeg)~\cite{mmseg2020}.

\noindent \textbf{Model Evaluation.} We utilize multiple metrics to determine the optimal model parameters during training and ablation studies. The primary evaluation metrics are kNN accuracy and linear probing performance on the fMoW-RGB dataset. For kNN we use $k=20$.

To evaluate the model's robustness to unseen resolutions and GSDs, we resize some of the classification datasets to scales of 100\%, 50\%, 25\%, and 12.5\% of their original resolution and compute the average kNN accuracy across these scales. We use the RESISC45, WHU-RS19, and SIRI-WHU datasets for this analysis due to their diverse GSDs and resolutions.

\section{Experiments}

In this section, we compare the quality of learned features between our approach and multiple MAE-based methods. We employ standard evaluation techniques, such as linear probing and kNN classification. Additionally, we assess our model’s performance on downstream tasks, including classification and segmentation.

\begin{table}[!htbp]
\centering
\footnotesize
\setlength{\tabcolsep}{0.25em}
\begin{tabular}{l|cccc}
\toprule
Dataset    & \footnotesize{SatDINO$_8$} & \footnotesize{SatDINO$_{16}$} & \footnotesize{Scale-MAE} & \footnotesize{SatMAE} \\
\midrule
EuroSAT    &   \textbf{87.72}   &   85.96   &   85.42   &   81.43 \\
RESISC45   &   \textbf{85.29}   &   82.32   &   79.96   &   65.96 \\
UC Merced  &   \textbf{94.82}   &   93.21   &   84.58   &   78.45 \\
WHU-RS19   &   \textbf{98.18}   &   97.82   &   89.32   &   86.41 \\
RS-C11     &   \textbf{96.91}   &   96.61   &   93.03   &   83.96 \\
SIRI-WHU   &   \textbf{91.82}   &   87.19   &   84.84   &   77.76 \\
\bottomrule                                
\end{tabular}
\caption{\textbf{Average kNN classification accuracy across multiple scales.} We report the mean kNN accuracy for SatDINO-Small and Scale-MAE across all evaluated datasets. SatDINO achieves consistently higher performance, surpassing Scale-MAE by up to 9\% in some cases.}
\label{tab:average_knn}
\end{table}

\begin{table*}[!t]
\centering
\footnotesize
\begin{tabular}{lllcc|cc|cc}
\toprule
                          &                 &           &    &    & \multicolumn{2}{c}{Linear probe.} & \multicolumn{2}{c}{Fine-tune} \\
                          \midrule
\multicolumn{1}{c}{Model} & Source          & Backbone  & Params.   & GFLOPs    & Top-1       & Top-5      & Top-1       & Top-5      \\
\midrule
SatMAE                    & SatMAE          & ViT-Large        & 303.11   & 119.37       & 65.9                 & -          & 77.8        & -          \\
SatMAE                    & Scale-MAE       & ViT-Large        & 303.11   & 119.37       & -                    & -          & 72.4        & 91.9       \\
SatMAE                    & Cross-MAE       & ViT-Large        & 303.11   & 119.37       & 67.8                 & 92.3       & -           & -          \\
SatMAE                    & Cross-MAE       & ViT-Base         & 85.65    & 33.73        & 65.1                 & 89.2       & -           & -          \\
\midrule
Scale-MAE                 & Scale-MAE       & ViT-Large        & 303.11   & 119.37       & -                    & -          & \textbf{77.9}        & 94.3       \\
Scale-MAE                 & Cross-MAE       & ViT-Base         & 85.65    & 33.73        & 67.3                 & 93.0       & -           & -          \\
\midrule
Cross-MAE                 & Cross-MAE       & ViT-Base         & 85.65    & 33.73        & 69.2                 & \textbf{95.4}       & -           & -          \\
\midrule
SatDINO                  & Ours            & ViT-Small${_{16}}$  & 21.59     & 8.54      & 72.75                & 92.38      & 77.62       & 94.11      \\
SatDINO                  & Ours            & ViT-Small${_8}$     & 21.37     & 33.56     & \textbf{73.53}       & 92.76      & 77.80       & \textbf{94.41}      \\
SatDINO                  & Ours            & ViT-Base            & 85.65     & 33.90     & \textbf{73.52}       & 92.43      & 77.29       & 94.04      \\
\bottomrule
\end{tabular}
\caption{\textbf{Comparison of linear probing and fine-tuning performance on the fMoW-RGB dataset.} Our model achieves the highest Top-1 accuracy in linear probing, demonstrating the strength of its learned representations. In fine-tuning, our model performs competitively with other state-of-the-art methods, achieving similar or better results while using a smaller architecture. Note that the setting for linear probing and fine-tuning may differ between the models.}
\label{tab:fmow_linprobe_finetune}
\end{table*}

\noindent \textbf{Training setup.} We pre-trained SatDINO-Small from scratch with patch sizes of 16 and 8. The model with a patch size of 16 was pre-trained for 200 epochs on the fMoW-RGB dataset using a batch size of 64 per GPU across 8 GPUs. The training was conducted with the AdamW~\cite{loshchilov2017decoupled} optimizer and a cosine learning rate scheduler with a 10 epoch warm-up. After the warm-up phase, the learning rate was 0.001. For the model with a patch size of 8, we used the same settings, except it was trained for 100 epochs with a batch size of 32.

\subsection{kNN Performance}  
To evaluate the quality and robustness of the learned features, we perform kNN classification on a variety of remote sensing datasets, using a frozen model. We extract features using the \textit{class token} as the representation. To further assess feature robustness, we evaluate the model across multiple scales.

For fair comparison, we used our fixed dataset splits and publicly available code and checkpoints from SatMAE and Scale-MAE to replicate their results. However, due to differences in dataset splits, our results may differ from those reported in the SatMAE and Scale-MAE papers. Unfortunately, we were unable to successfully reproduce the results of Cross-Scale MAE using their provided code and checkpoints.

A direct comparison with SatMAE and Scale-MAE is presented in Figure \ref{fig:knn}, where our model consistently outperforms SatMAE and Scale-MAE across almost all datasets and scales. For a more comprehensive comparison, Table \ref{tab:average_knn} shows the average kNN performance across scales for both models. SatDINO outperforms Scale-MAE on all datasets, with improvements exceeding 9\% in some cases. This shows that SatDINO learns stronger and more robust features than previous MAE-based approaches.

\subsection{Linear Probing and Fine-Tuning}
To further assess the robustness of learned features and the adaptability of our model, we evaluate its performance using linear probing and fine-tuning on classification datasets.

\begin{table}[!htbp]
\centering
\footnotesize
\begin{tabular}{l|ccc}
\toprule
Dataset   &  Small$_{16}$   &Small$_8$      & Base  \\
\midrule
EuroSAT   &  98.69          &98.76          & \textbf{98.83} \\
RESISC45  &  95.68          &95.16          & \textbf{96.05} \\
UC Merced &  98.33          &\textbf{98.81} & 98.57          \\
WHU-RS19  &  \textbf{98.54} &98.06          & 97.57          \\
RS-C11    &  \textbf{98.01} &96.81          & 96.02          \\
SIRI-WHU  &  \textbf{98.54} &97.08          & 97.08          \\
\bottomrule
\end{tabular}
\caption{\textbf{SatDINO fine-tuning classification results.} All three variants of the SatDINO model achieve consistently strong classification performance across multiple datasets.}
\label{tab:classification_finetune}
\end{table}

\begin{table}[!htbp]
\centering
\footnotesize
\setlength{\tabcolsep}{0.25em}
\begin{tabular}{ll|cc|cc|cc}
\toprule
          &                              & \multicolumn{2}{c}{Potsdam} & \multicolumn{2}{c}{Vaihingen} & \multicolumn{2}{c}{LoveDA} \\
\midrule
Model     & Backbone                     & 224$^2$          & 512$^2$          & 224$^2$           & 512$^2$           & 224$^2$          & 512$^2$         \\
\midrule
SatMAE    & ViT-Large                    & 67.88        & 70.39        & 64,81         & 69.13         &              46.28&             52.28\\
Scale-MAE & ViT-Large                    & 69.74        & \textbf{72.21}        &               67.97&               \textbf{71.65}& \textbf{49.37}        & \textbf{53.70}       \\
\midrule
SatDINO   & ViT-Small$_{16}$             & 67.93        & 71.80        &               63.38&               68.32& 44.77        & 49.65       \\
SatDINO   & ViT-Small$_8$                & \textbf{70.71}        & 71.45        &               \textbf{68.69}&               67.71& 47.53        & 50.20       \\
SatDINO   & ViT-Base                     & 67.65        & 71.63        &               64.85&               69.37& 44.25        & 50.08       \\
\bottomrule
\end{tabular}
\caption{\textbf{Comparison of semantic segmentation performance across multiple datasets and image scales.} All results are reported in terms of mean Intersection over Union (mIoU). The smaller image size (224$^2$) corresponds to the resolution at which most models were trained, while the larger size (512$^2$) reflects the default resolution of the datasets used.}
\label{tab:segmentation}
\end{table}

For linear probing, we freeze the entire model and add a single linear layer on top of the \textit{class token}. The model is then trained for 25 epochs using a learning rate of 0.001. For fine-tuning, we follow the same setup but do not freeze the model, allowing all layers to be updated. In this case, we use a learning rate of 0.00001.

\begin{table*}[!ht]
    \begin{subtable}[t]{0.55\textwidth}
        \centering
        \footnotesize
        \setlength{\tabcolsep}{0.35em}
        \begin{tabular}{cl|ccc|ccc}
        \toprule
        \begin{tabular}[c]{@{}l@{}}Temp.\\ aug.\end{tabular} & Aug. & kNN            & top-1          & top-5  & \scriptsize{RESISC45} & \scriptsize{WHU-RS19} &  \scriptsize{SIRI-WHU}  \\
        \midrule
        \xmark       & default       & \textbf{68.89} & \textbf{71.22} & \textbf{91.87}           &  82.17                & \textbf{97.69}       & 87.55       \\
        \xmark       & none          & 63.35          & 65.98          & 88.81                    &  80.30                & 95.27       & \textbf{88.28}       \\
        \xmark       & soft          & 68.60          & 71.03          & 91.49                    &  81.20                & 97.21       & 84.69       \\
        \xmark       & mid           & \textbf{68.96} & \textbf{71.28} & \textbf{91.80}           &  \textbf{82.44}       & 97.09       & 87.92       \\
        \midrule
        \cmark       & default       & 67.69          & 71.24          & \textbf{91.96}  &  80.17                & 97.21       & 85.63       \\
        \cmark       & none          & 67.19          & 69.85          & 91.15                    &  80.10                & 96.00       & \textbf{86.93}       \\
        \cmark       & soft          & 67.19          & 70.96          & 91.60                    &  79.32                & 97.33       & 83.23       \\
        \cmark       & mid           & \textbf{68.28} & \textbf{71.72} & \textbf{92.01}           &  \textbf{81.04}       & \textbf{97.45}       & 86.20       \\ 
        \bottomrule
        \end{tabular}
        \caption{Effect of augmentations, temporal augmentations and their intensity.}
        \label{tab:ablation_augs}
    \end{subtable}
    \hspace{\fill}
    \begin{subtable}[t]{0.44\textwidth}
        \centering
        \footnotesize
        \setlength{\tabcolsep}{0.40em}
        \begin{tabular}{@{}cll|cll@{}}
        \toprule
        \begin{tabular}[c]{@{}l@{}}Temp.\\ aug.\end{tabular} & \begin{tabular}[c]{@{}l@{}}Local\\ scale\end{tabular} & \begin{tabular}[c]{@{}l@{}}Global\\ scale\end{tabular} & kNN            & top-1          & top-5          \\
        \midrule
        \xmark                             & $\left[ 5, 25 \right]$                                & $\left[ 25, 100 \right]$                               & 68.89          & \textbf{71.22} & \textbf{91.87} \\
        \xmark                             & $\left[ 5, 50 \right]$                                & $\left[ 50, 100 \right]$                               & \textbf{69.18} & 71.11          & 91.52          \\
        \xmark                             & $\left[ 25, 50 \right]$                               & $\left[ 50, 100 \right]$                               & 66.25          & 67.89          & 89.74          \\
        \midrule
        \cmark                             & $\left[ 5, 25 \right]$                                & $\left[ 25, 100 \right]$                               & \textbf{67.69} & \textbf{71.14} & \textbf{91.96} \\
        \cmark                             & $\left[ 5, 50 \right]$                                & $\left[ 50, 100 \right]$                               & 67.27          & 70.90          & 91.69          \\
        \cmark                             & $\left[ 25, 50 \right]$                               & $\left[ 50, 100 \right]$                               & 64.57          & 67.28          & 89.82         \\
        \bottomrule
        \end{tabular}
        \caption{Comparison of different scales and temporal augmentations.}
        \label{tab:ablation_scale}
    \end{subtable}
\caption{\textbf{Ablation on augmentation strength and application of temporal augmentations.} We evaluate the impact of \textit{soft} and \textit{mid} augmentations by scaling their application probabilities to 0.25$\times$ and 0.75$\times$ of the \textit{default}. Additionally, we examine temporal augmentations, where different images of the same area over time are used for each global and local views.}
\label{tab:table1}
\end{table*}


Table \ref{tab:fmow_linprobe_finetune} presents a comparison of linear probing and fine-tuning performance on the fMoW-RGB dataset. We include results from multiple sources. Our model outperforms all others in linear probing Top-1 classification accuracy and achieves comparable performance in other tasks. Notably, we achieve these results with a smaller model, demonstrating the efficiency of our approach.

We also fine-tuned our model on the classification datasets we used for kNN. We trained for 25 epochs with a learning rate of 0.001. Results are in Table~\ref{tab:classification_finetune}.

\subsection{Segmentation Results}
Finally, we evaluate SatDINO on semantic segmentation tasks. We use pre-trained ViT models as the backbone for UperNet~\cite{xiao2018unified} and compare SatDINO against SatMAE and Scale-MAE across three segmentation datasets at two input scales.

\noindent \textbf{Training setup.} We train the models using the MMSeg~\cite{mmseg2020} framework for 60k iterations with a cosine learning rate scheduler and a warm-up phase. After warm-up, the learning rate is set to 0.00006. We use the AdamW optimizer with a batch size of 8.

Table \ref{tab:segmentation} presents the semantic segmentation results in terms of mIoU. SatDINO demonstrates competitive performance across multiple settings. The ViT-Small$_8$ variant performs particularly well at lower resolutions, achieving the highest mIoU on Potsdam at 224$^2$ (70.71) and Vaihingen at 224$^2$ (68.69), outperforming both SatMAE and Scale-MAE at these settings. The ViT-Small$_{16}$ and ViT-Base variants perform well at higher resolutions (512$^2$) but do not surpass Scale-MAE. Notably, Scale-MAE utilizes a ViT-Large model, which has significantly more parameters than the SatDINO.

\subsection{Discussion}
SatDINO performs strongly in kNN and linear probing tasks, making it ideal for embedding-based applications. Its compact size and low computational cost enable efficient processing of large image datasets, which is especially useful for tasks like image matching that require compressing images into vector representations.

In segmentation tasks, SatDINO performs well at lower resolutions but falls behind MAE-based models at higher resolutions. This performance gap may be attributed to the larger ViT models used in MAE-based approaches, which have a greater capacity for learning during fine-tuning. This could be addressed by using the ViT-Large variant of SatDINO. However, our experiments showed signs of overfitting when training larger models, suggesting that DINO requires more data for effective training at scale. Further research is needed to explore this limitation.

\section{Ablation Study} \label{ablation}
 
In the ablation study, we show how individual components presented in Section~\ref{methodology} influence the overall performance. We use ViT-Small in all experiments and always evaluate the teacher model. We use kNN accuracy and linear-probing top-1 and top-5 accuracy on fMoW-RGB as main metrics and if needed, we also show average kNN accuracy across multiple scales on downstream classification datasets.

\noindent \textbf{Training setup.} Our baseline training setup is run for 100 epochs with a learning rate of 0.001, cosine learning rate scheduler, AdamW optimizer, and batch size 64 on 8 GPUs. We also use standard DINO augmentations with 8 local views and 2 global views, the local scale of the views is $\left[ 5, 25 \right]$ and the global scale is $\left[ 25, 100 \right]$.

\noindent \textbf{Evaluation setup.} Liner probing is achieved by adding one linear layer on top of the frozen model. We are training for 25 epochs with learning rare 0.0001 with AdamW optimizer and batch size 64.

\begin{table*}[!htbp]
    \centering
    \begin{subtable}[t]{0.30\textwidth}
        \centering
        \footnotesize
        \begin{tabular}{c|ccc}
        \toprule
        \begin{tabular}[c]{@{}l@{}}Views\\ number\end{tabular} & kNN            & top-1          & top-5          \\
        \midrule
        2          & 65.27          & 67.31          & 89.70          \\
        4          & 67.60          & 69.83          & 90.99          \\
        6          & 68.92          & 71.09          & 91.61          \\
        8          & 69.02          & 71.23          & 91.88          \\
        10          & \textbf{69.42} & \textbf{72.03} & \textbf{92.18} \\
        12          & 69.29          & \textbf{71.99} & \textbf{92.18}          \\
        \bottomrule
        \end{tabular}
        \caption{Effect of a number of local views.}
        \label{tab:ablation_num_views}
    \end{subtable}
    \hspace{\fill}
    \begin{subtable}[t]{0.66\textwidth}
        \centering
        \footnotesize
        \begin{tabular}{ll|ccc|ccc}
        \toprule
        \begin{tabular}[c]{@{}l@{}}Local\\ scale\end{tabular} & \begin{tabular}[c]{@{}l@{}}Global\\ scale\end{tabular} & kNN    & top-1   & top-5   & \footnotesize{RESISC45} & \footnotesize{WHU-RS19} &  \footnotesize{SIRI-WHU}  \\
        \midrule
        $\left[ 5, 25 \right]$    & $\left[ 25, 100 \right]$    & \textbf{69.34} & \textbf{71.87}   & \textbf{92.15} &  82.21	&	97.45	&   \textbf{87.40}   \\
        $\left[ 5, 50 \right]$    & $\left[ 50, 100 \right]$    & 69.05          & 71.17            & 91.52          &  82.29	&	98.06	&	87.08   \\
        $\left[ 25, 50 \right]$    & $\left[ 50, 100 \right]$    & 66.80          & 68.45           & 89.96          &  82.13	&	98.30	&	86.20   \\
        $\left[ 5, 50 \right]$    & $\left[ 25, 100 \right]$    & 68.44          & 70.86            & 91.50          &  82.70	&	98.06	&	86.20   \\
        $\left[ 5, 75 \right]$    & $\left[ 25, 100 \right]$    & 68.38          & 71.15            & 91.34          &  82.79	&	98.06	&	86.51   \\
        $\left[ 5, 100 \right]$    & $\left[ 25, 100 \right]$    & 67.30          & 69.94           & 91.11          &  \textbf{83.25}	&	\textbf{98.42}	&	86.82   \\
        \bottomrule
        \end{tabular}
        \caption{Effect of uniform scale selection strategy.}
        \label{tab:ablation_uniform}
    \end{subtable}    
    
    \bigskip
    
    \begin{subtable}[t]{0.66\textwidth}
        \centering
        \footnotesize
        \begin{tabular}{cc|ccc|ccc}
        \toprule
        \begin{tabular}[c]{@{}l@{}}Temp.\\ aug.\end{tabular} & \begin{tabular}[c]{@{}l@{}}Views\\ number\end{tabular} & kNN            & top-1          & top-5  & \footnotesize{RESISC45} & \footnotesize{WHU-RS19} &  \footnotesize{SIRI-WHU}  \\
        \midrule
        \xmark        & 4        & 67.75          & 69.92          & 91.24           &  77.06	&	96.72	&  84.90    \\
        \xmark        & 6        & \textbf{69.07} & \textbf{71.62} & \textbf{91.89}  &  \textbf{80.99}	&	\textbf{96.84	}&  \textbf{86.20}   \\
        \midrule
        \cmark        & 4        & 66.62          & 70.26          & 91.48           &  76.02	&	95.87	&  83.18    \\
        \cmark        & 6        & \textbf{67.46} & \textbf{71.41} & \textbf{92.07}  &  \textbf{79.95}	&	\textbf{97.09}	&  \textbf{85.10}   \\
        \bottomrule
        \end{tabular}
        \caption{Effect of local view variable size strategy.}
        \label{tab:ablation_variable_size}
    \end{subtable}
    \hspace{\fill}
    \begin{subtable}[t]{0.30\textwidth}
        \centering
        \footnotesize
        \begin{tabular}{c|ccc}
        \toprule
        \begin{tabular}[c]{@{}l@{}}Aspect\\ ratio\end{tabular} & kNN            & top-1          & top-5          \\
        \midrule
        $\left[ \frac{3}{4}, \frac{4}{3} \right]$          & \textbf{68.98}          & \textbf{71.55}          & \textbf{91.91}          \\
        $\left[ 1, 1 \right]$                              & 65.77          & 70.36          & 91.32          \\
        $\left[ \frac{1}{2}, \frac{3}{2} \right]$          & \textbf{69.06}          & 71.39          & 91.69          \\
        $\left[ \frac{1}{2}, 2 \right]$                    & 68.83          & 71.21          & 91.79          \\
        \bottomrule
        \end{tabular}
        \caption{Local view crop aspect ratio.}
        \label{tab:ablation_aspect_ratio}
    \end{subtable}
\caption{\textbf{Ablation on local views and sampling strategy.} We show how performance scales with an increasing number of views. We evaluate two sampling strategies: our method, which samples local views uniformly to enhance GSD diversity, and the approach from \cite{Wanyan_2024_CVPR}, where each local view is resized to a different size. Additionally, we briefly examine the effect of the aspect ratio of the cropped views.}
\label{tab:table1}
\end{table*}

\subsection{Augmentations} 
We begin by evaluating the impact of augmentations applied to local and global views. Table \ref{tab:ablation_augs} presents a comparison of augmentations with different intensities, both with and without the use of \textit{temporal augmentations}. The \textit{default} and \textit{mid} augmentation intensity demonstrated the best performance on the fMoW-RGB dataset and downstream classification tasks. Results on fMoW-RGB with and without \textit{temporal augmentations} were very similar, with \textit{temporal augmentations} occasionally leading to slight improvements. However, performance on downstream classification tasks consistently decreased when \textit{temporal augmentations} were used. Therefore, we chose to use \textit{default} augmentations and exclude \textit{temporal augmentations}, as they add unnecessary complexity without clear benefits.

The size of the local view can dramatically affect the class of the image. Table \ref{tab:ablation_scale} illustrates how increasing the size of the local view impacts the performance. All models in this experiment were trained using \textit{default} augmentations. Interestingly, the smallest local view size achieved the best results. This indicates that the model can infer the class even from highly distorted images. This shows the importance of designing a pretraining task complex enough for the model to learn strong features. 


\subsection{Local views} 
Next, we evaluate the impact of local views and alternative methods for sampling them. We begin with the default setting, where local crops are sampled randomly from the default scale range. Table \ref{tab:ablation_num_views} demonstrates how the number of local views affects performance. In our experiments, performance peaked at 10 local views.

We then tested two alternative sampling methods. The first method, proposed in \cite{Wanyan_2024_CVPR}, involves randomly cropping local scales and resizing them to different sizes. We evaluated this method using 4 and 6 local views, along with temporal augmentations, as described in \cite{Wanyan_2024_CVPR}. The results in Table \ref{tab:ablation_variable_size} show that increasing the number of local views improves performance, consistent with the previous case. However, the inclusion of temporal augmentations led to worse results, mainly on the fMoW-RGB dataset.

The second sampling method we tested is the one we proposed. Instead of randomly sampling from the full range for each view, we divided the range into uniform sub-ranges based on the number of views and then randomly sampled one view from each sub-range. We tested several ranges with 8 local views, and the results are shown in Table \ref{tab:ablation_uniform}. Using the default range (local: $\left[5, 25\right]$, global: $\left[25, 100\right]$) produced the best results on the fMoW-RGB dataset. However, expanding the local view range, improved performance on downstream tasks.

\begin{table}[!t]
    \centering
    \begin{subtable}[t]{0.47\textwidth}
        \centering
        \scriptsize
        \setlength{\tabcolsep}{0.35em}
        \begin{tabular}{l|ccc|ccc}
        \toprule
        \begin{tabular}[c]{@{}l@{}}GSD Loss\\ weight\end{tabular} & kNN   & top-1   & top-5  & \footnotesize{RESISC45} & \footnotesize{WHU-RS19} & \footnotesize{SIRI-WHU} \\
        \midrule
        0             & 68.86          & 71.22          & 91.71                   & 82.21          & 97.45          & 87.45          \\
        0.0001        & \textbf{68.91} & \textbf{71.41}          & \textbf{91.77} & \textbf{82.56} & 97.45          & 87.50          \\
        0.001         & \textbf{68.91} & 71.33          & 91.63                   & 82.42          & \textbf{97.69} & 87.19          \\
        0.01          & 68.67          & 71.17          & \textbf{91.77}          & 82.38          & 97.45          & \textbf{87.86} \\
        0.1           & 68.84          & 71.30          & 91.75                   & \textbf{82.55} & \textbf{98.18} & 87.40          \\

        \bottomrule
        \end{tabular}
        \caption{GSD encoding loss weight.}
        \label{tab:ablation_gsd_weight}
    \end{subtable}

        \bigskip
    
    \begin{subtable}[t]{0.47\textwidth}
        \centering
        \scriptsize
        \setlength{\tabcolsep}{0.30em}
        \begin{tabular}{ll|ccc|ccc}
        \toprule
        \begin{tabular}[c]{@{}l@{}}Local\\ scale\end{tabular} & \begin{tabular}[c]{@{}l@{}}Global\\ scale\end{tabular} & kNN            & top-1          & top-5          & \scriptsize{RESISC45} & \scriptsize{WHU-RS19} & \scriptsize{SIRI-WHU} \\
        \midrule
        $\left[5, 25\right]$         & $\left[25, 100\right]$         & \textbf{68.83} & \textbf{71.22} & \textbf{91.67} & 82.67                   & 97.69                   & 87.50                   \\
        $\left[5, 50\right]$         & $\left[50, 100\right]$         & 68.20          & 70.52          & 91.47          & 82.37                   & 97.69                   & 86.04                   \\
        $\left[25, 50\right]$        & $\left[50, 100\right]$         & 66.87          & 68.11          & 89.84          & 82.42                   & \textbf{98.18}          & 86.67                   \\
        $\left[5, 50\right]$         & $\left[25, 100\right]$         & 68.29          & 70.81          & 91.63          & 83.37                   & 97.94                   & 87.40          \\
        $\left[5, 75\right]$         & $\left[25, 100\right]$         & 67.95          & 70.56          & 91.43          & \textbf{83.41}          & 97.94                   & \textbf{87.55}          \\
        $\left[5, 100\right]$        & $\left[25, 100\right]$         & 66.86          & 69.92          & 91.01          & 83.13                   & 97.82                   & 86.61                   \\
        \bottomrule
        \end{tabular}
        \caption{GSD encoding with uniform local scale.}
        \label{tab:ablation_gsd_scale}
    \end{subtable}

    \bigskip
    
    \begin{subtable}[t]{0.36\textwidth}
        \centering
        \scriptsize
        \setlength{\tabcolsep}{0.35em}
        \begin{tabular}{ll|ccc}
        \toprule
        \begin{tabular}[c]{@{}l@{}}Temp.\\ aug\end{tabular} & Aug. & kNN            & top-1          & top-5          \\
        \midrule
        \xmark           & default       & \textbf{68.81} & \textbf{71.20}          & \textbf{91.80}          \\
        \xmark           & mid           & 68.59          & 70.80          & 91.55          \\
        \midrule
        \cmark           & default       & 67.62          & \textbf{71.41} & \textbf{91.98} \\
        \cmark           & mid           & \textbf{67.94}          & 71.19          & 91.76          \\
        \bottomrule
        \end{tabular}
        \caption{GSD encoding with augmentations.}
        \label{tab:ablation_gsd_augmentations}
    \end{subtable}

\caption{\textbf{Ablation study on GSD encoding.} We examine the impact of different GSD loss weights on model performance. Additionally, we assess the effect of GSD encoding on various augmentations, as well as its effect when combined with uniform local view sampling.}
\label{tab:table1}
\end{table}

When comparing the two methods, our proposed method achieved better results on both the fMoW-RGB dataset and downstream classification tasks. Additionally, the first method has a computational disadvantage, as it requires larger local views. It also makes increasing the number of views less straightforward. In contrast, both our method and the default approach enable easy scalability by simply increasing the number of views. When comparing our method to the default random sampling, our method slightly outperforms the default approach.

In the default DINO setting, views are cropped as rectangles with an aspect ratio in the range $\left[\frac{3}{4}, \frac{4}{3}\right]$, and then resized to squares. This resizing introduces distortion, which we found to be very useful for training. Table \ref{tab:ablation_aspect_ratio} shows the effect of using different aspect ratio ranges. When views are cropped as squares ($\left[1, 1\right]$), mainly kNN performance drops, this suggests that the model learns less informative features without this augmentation. 


\subsection{GSD encoding}
Lastly, we evaluate the impact of GSD encoding on training. We began by testing different values for the GSD loss weight $\gamma$. The results, summarized in Table \ref{tab:ablation_gsd_weight}, indicate that decreasing the weight improved performance on the fMoW-RGB dataset, while a larger weight improved performance on downstream classification tasks. To balance these trade-offs and emphasize GSD encoding, we selected $\gamma = 0.1$ in subsequent experiments.

To ensure consistency, we repeated some experiments involving augmentations and \textit{temporal augmentations}. The results corresponded with previous findings and again confirmed that \textit{temporal augmentations} do not improve performance. These results are presented in Table \ref{tab:ablation_gsd_augmentations}.

Finally, we tested GSD encoding in combination with uniform local view sampling, as both approaches are concerned with GSD. The results, shown in Table \ref{tab:ablation_gsd_scale}, were consistent with those in Table \ref{tab:ablation_scale}. Except that performance on the fMoW-RGB dataset slightly decreased, while the results for downstream classification tasks slightly improved.

\subsection{Summary}
We summarize the main results in Table \ref{tab:ablation_summary}. The experiments were conducted using a ViT-Small model with 10 local views sampled from the range $\left[5, 25\right]$. For experiments involving GSD encoding, the GSD loss weight was set to $\gamma = 0.1$.

When uniform sampling was used independently, performance on the fMoW-RGB dataset improved, but performance on downstream classification tasks decreased. In contrast, when only GSD encoding was applied, the fMoW-RGB performance declined while downstream classification performance improved. Finally, when both uniform sampling and GSD encoding were used together, almost all results showed improvement.

\begin{table}[!htbp]
\centering
\footnotesize
\setlength{\tabcolsep}{0.35em}
\begin{tabular}{cc|ccc|ccc}
\toprule
\begin{tabular}[c]{@{}l@{}}\scriptsize{Uni.}\\ \scriptsize{scale}\end{tabular} & \begin{tabular}[c]{@{}l@{}}\scriptsize{GSD}\\ \scriptsize{enc.}\end{tabular} & \scriptsize{kNN}  & \scriptsize{top-1}  & \scriptsize{top-5}  & \scriptsize{RESISC45} & \scriptsize{WHU-RS19} & \scriptsize{SIRI-WHU} \\
\midrule
\xmark       & \xmark       & 69.16       & 72.31       & 92.19       & 82.12       & 97.57       & 87.29       \\
\midrule
\cmark       & \xmark       & \textcolor{ForestGreen}{0.13}          & \textcolor{ForestGreen}{0.11}            & \textcolor{ForestGreen}{0.11}       & \textcolor{BrickRed}{-0.41}       & \textcolor{BrickRed}{-0.24}       & \textcolor{BrickRed}{-0.31}       \\
\xmark       & \cmark       & \textcolor{BrickRed}{-0.18}            & \textcolor{BrickRed}{-0.01}              & \textcolor{ForestGreen}{0.10}       & \textcolor{ForestGreen}{\textbf{0.30}}       & \textcolor{ForestGreen}{0.00}       & \textcolor{ForestGreen}{\textbf{0.16}}       \\
\cmark       & \cmark       & \textcolor{ForestGreen}{\textbf{0.31}} & \textcolor{ForestGreen}{\textbf{0.44}}   & \textcolor{ForestGreen}{\textbf{0.19}}       & \textcolor{ForestGreen}{0.21}       & \textcolor{ForestGreen}{\textbf{0.24}}       & \textcolor{BrickRed}{-0.10}       \\
\bottomrule
\end{tabular}
\caption{\textbf{SatDINO ablation summary.} We show the impact of uniform local scale sampling and GSD encoding on fMoW performance and kNN downstream classification. Uniform scale sampling enhances performance on the fMoW pretraining dataset, while GSD encoding improves downstream task performance. When combined, the performance improves across most metrics.}
\label{tab:ablation_summary}
\end{table}

\section{Conclusion}
In this work, we introduced a novel self-supervised method for remote sensing - SatDINO. We demonstrated that SatDINO significantly improves feature extraction and robustness across various datasets. In extensive evaluations of classification and segmentation tasks, SatDINO achieved competitive results and overcame multiple MAE-based approaches. Additionally, we performed an extensive ablation study in which we evaluated individual SatDINO components and also possible training setups.

In future work, we would like to focus on pretraining larger models. We anticipate that larger models will require more data to be effectively pretrained with DINO. Our experiments have shown how important it is for the pretraining task to be sufficiently complex. One possible research direction is to explore how individual augmentations affect pretraining and, for example, progressively increase the complexity of the pretraining task.


\bibliographystyle{ieeetr}
\bibliography{sample}

\end{document}